\documentclass[journal]{IEEEtran}
\usepackage[pdftex]{graphicx}
\usepackage{acronym}
\usepackage{multirow}
\usepackage{amsmath, amsthm, amssymb, amsfonts, mathtools}
\usepackage{placeins}
\usepackage{algorithm} 
\usepackage{algpseudocode} 
\usepackage{url}
\usepackage{hhline}
\usepackage[stable]{footmisc}
\renewcommand{\baselinestretch}{1.0}\normalsize

\usepackage{tikz}
\usepackage{pgfplots}
\usepackage{times}
\usetikzlibrary{calc}
\usetikzlibrary{backgrounds}
\usepackage{soul}
\usepackage{authblk}

\usepackage{xcolor}

\providecommand{\keywords}[1]
{
  \textbf{\textit{Keywords---}} #1
}

\title{One-Shot Federated Learning with Neuromorphic Processors}
\author[1,*]{Kenneth Stewart} 
\author[1,*]{Yanqi Gu}
\affil[1]{Department of Computer Science, University of California, Irvine, Irvine, CA, United States}
\affil[*]{Both authors contributed equally to this work}

\begin{document}

\maketitle

\section{Abstract}
Being very low power, the use  of neuromorphic processors in mobile devices to solve machine learning problems is a promising alternative to traditional Von Neumann processors. Federated Learning enables entities such as mobile devices to collaboratively learn a shared model while keeping their training data local.
Additionally, federated learning is a secure way of learning because only the model weights need to be shared between models, keeping the data private.
Here we demonstrate the efficacy of federated learning in neuromorphic processors. 
Neuromorphic processors benefit from the collaborative learning, achieving state of the art accuracy on a one-shot learning gesture recognition task across individual processor models while preserving local data privacy.

\hspace{8pt}

\keywords{\textbf{neuromorphic computing, one-shot learning, federated learning, data privacy}}

\section{Introduction}
Recent progress in neuromorphic processors, sensors, and Spiking Neural Network (SNN) learning algorithms has created an energy efficient alternative to the widely used GPUs and Artificial Neural Network algorithms. 
Because of their energy efficiency, neuromorphic processors and sensors are practical for mobile devices such as phones and robots with constrained power budgets.
While recent work has demonstrated the success of on chip learning on neuromorphic processors for classification tasks \cite{Stewart_etal20_on-cfew-} \cite{tang2019spiking}, being used in mobile devices gives the opportunity to access additional data through crowd sourcing that could improve the generalizability of the SNN models used like their ANN counterparts.
However data obtained through crowd sourcing may contain sensitive information that users will want to keep private.
Here we combine a state of the art Surrogate Gradient Online Error-triggered Learning (SOEL) \cite{stewart2020online} SNN learning algorithm with the advanced privacy preserving federated averaging algorithm, demonstrating state of the art one-shot learning accuracy on a gesture recognition task that keeps data private.

\section{Background}
\subsection{Federated learning}
Traditionally, in a standard centralized training process, the learning model is trained by the central server who collects the training data from local datasets, e.g. data from mobile devices or sensors. However, this could cause privacy issues because data owners have to share their raw data with the central server, and the data itself may contain sensitive information that needs to be kept private. To overcome this shortcoming, Federated Learning was proposed by Google \cite{googleFed} to allow clients and the server to share a common learning model without uploading clients' private training data. Its backbone is an algorithm called Federated Averaging, which is shown in Algorithm 1. The idea is to distribute and run the learning algorithms locally on the client's sides instead of the central server side. The local models are periodically synchronized with the central model. For example, in
Federated Learning, clients send their model updates (weight updates) to the central server, and then the server summarizes the collected weights into a common model and returns this model to clients. The process repeats until the model converges. Recently Federated Learning has become a popular privacy solution when considering the potential applications of training shared model from crowd sourced data, including input text prediction \cite{fedtext}, mobile ad recommendation,  or medical data analysis \cite{covidfed} over the confidential data collected from many different hospitals.

\begin{algorithm}
\begin{algorithmic}

   \State \textbf{Server}:
   \State Server initializes common model $w_0$    
    \For {$t = 1$, $E$}   
        \For {each client $k$ in $K$}   
            \State $w_{t} = w_{t-1} +$\textbf{Client}$_k(w_{t-1}^k)$   
        \EndFor   
        \State $w_{t} = w_{t}/K$   
    \EndFor   
    
    \State \textbf{Output}: Global model $w_t$
    \\
    \State \textbf{Client$_k(w_{t-1}^k)$:}
    \State $W_{e}^{k} =$ SGD$(D_k,w_{t-1}^{k},T)$
    \State \textbf{Output}: Model update $w_t^k - w_{t-1}^k$
\end{algorithmic}

\label{alg:FedAvg}
\caption{Federated Averaging. Input: $E$ is the number of epochs on the server side, $T$ is number of local epochs on client side, $W$ are the weights, $D$ are the training datasets, $K$ clients are indexed by $k$. Local models update via stochastic gradient descent(SGD)}
\end{algorithm} 

\subsection{Neuromorphic computing}
Neuromorphic computing platforms offer an energy-efficient alternative to perform training and inference in neural networks while being suitable for power-constrained applications such as mobile systems \cite{hwu2017complete}.
Neuromorphic systems mimic the brain's event-driven dynamics, distributed architecture and massive parallelism to overcome the limitations of conventional von Neumann computing architectures \cite{Indiveri_etal11_neursili}. 
Neuromorphic hardware equipped with synaptic plasticity capability can perform training and inference online, using local information \cite{Chicca_etal13_neurelec,Davies_etal18_loihneur}, making them particularly interesting for problems requiring fast adaptation to new data.
In previous work, \cite{Imam2020smell} showed fast adaptation to various odourants and \cite{stewart2020online} showed fast adaptation to mid-air gestures.

The key contribution of this work is showing how federated learning can be used to further improve one-shot learning accuracy using surrogate gradient learning on neuromorphic processors. 

\subsubsection{Surrogate Gradient Online Error-triggered Learning with the Loihi Neuromorphic Research Chip}
A number of recent methods for training SNNs using gradient descent have recently emerged. 
The Surrogate Gradient Online Error-triggered Learning (SOEL) learning rule has shown to be successful on classification tasks when used on the Intel Loihi neuromorphic processor. 
The Intel Loihi has a plasticity processor that can adjust synaptic weights via an  learning rule expressed as a finite difference equation with respect to a synaptic state variable that follows a sum-of-products form shown here \cite{Davies_etal18_loihneur}:

\begin{equation}\label{eq:loihi_rule}
    W_{ij}[t+1] = W_{ij}[t]+\sum_k C_{k}\prod_l F_{kl}[t],
\end{equation}

where $W_{ij}$ is the synaptic weight variable defined for the destination-source neuron pair being updated; $C_k$ is a scaling constant; and $F_{kl}[t]$ may be programmed to represent various state variables, including pre-synaptic spikes or traces, post-synaptic spikes or traces, where traces are represented as first-order linear filters. 
The weights are stochastically rounded according to the programmed weight precision. 
Traces are stochastically rounded to 7-bits of  precision. 
SOEL maps a surrogate gradient (SG) based \cite{Zenke_Ganguli17_supesupe} \cite{Neftci_etal19_surrgrad} three-factor learning rule to these dynamics on the Intel Loihi chip adapted to overcome limitations in neuromorphic hardware such as locality and bit precision. 
Three factor rules include a pre-synaptic factor, a post-synaptic factor and an external error signal. 
The SOEL learning rule can be written in the following three-factor form:

\begin{equation}\label{eq:eos_cont}
\nabla_{W_{ij}} \mathcal{L} (S) =
- (Y_i-S_i) 
\sigma'(U_i) 
P_j.
\end{equation}

where $Y_i$ is the target, $S_i$ is the number of post-synaptic spikes by neuron i, $\sigma'(U_i)$ is the derivative of the post-synaptic neuron's membrane state which must be implemented on the Loihi as a box function using a piecewise SG function where $\sigma'(U_i) \in \{0,1\}$, and $P_j$ is a subtraction of two pre-synaptic traces which creates a second order kernel.
The rule can be written compactly as follows:

\begin{equation}\label{eq:seol}
\nabla_{W_{ij}} \mathcal{L} (S^{N}) 
\propto
- E_i
B_i 
P_j.
\end{equation}

Since the post-synaptic trace is not necessary for the SG rule, SOEL writes the error on the same register used for the post-synaptic trace. This enables the error value to be available in the plasticity processor for learning.
On the Loihi, post-traces can only be positive but errors can be both positive and negative.
This problem is solved by offsetting the weight updates with a constant term $C$.

\begin{equation}\label{eq:eos_err}
    E_i[t] = \begin{cases} 
                C + {err}_i[t],\text{ if $err_i > \theta$ or $< -\theta$}\\
                C, \text{otherwise}
                \end{cases}
\end{equation}
 where $err_i[t] = Y-\bar{S}_i = \sum_{t-T}^T S_i[t]$ is the error calculated at timestep t which is the target $Y$ minus the number of post-synaptic spikes by neuron $i$ in the last $T$ time steps, and $\theta$ is an error threshold. 

The full learning rule can then be expressed as:
\begin{equation}
    W_{ij}= W_{ij} + \eta (E_i-C) P_j,
    \label{eq:LR}
\end{equation}

where $W_{ij}$ is the synapse from pre-synaptic neuron j to post-synaptic neuron i, $\eta$ is the learning rate, $E_i$ is the error, and $P_j$ is the pre-synaptic trace.
The learning rule can be implemented in Intel Loihi as:
\begin{equation}
    \begin{split}
    X^1_j[t+1] &= \alpha^1 X^1_j[t] + S^1_j,\\
    X^2_j[t+1] &= \alpha^2 X^2_j[t] + S^2_j,\\
    Y_i[t] &= E_i[t],\\
    \Delta W_{ij} &= \eta (X^2_j[t]-X^1_j[t])(Y_i[t] - C) .
    \end{split}
\end{equation}
Here, $X^2$ and $X^1$ are pre-synaptic trace variables available in the Loihi whose subtraction in the third equation yields the second order kernel equivalent to $P_j$ in Eq. (2). 
\begin{equation}
P_j [t] \propto (X^2_j[t]-X^1_j[t]).
\end{equation}
A Loihi Lakemont core computes the spike count $\bar{S}$ and evaluates $err_i$ at regular intervals $T$. If the error exceeds the threshold $\theta$, the post-synaptic trace value in the plasticity processor, $Y$, is written with the error $E_i$ and a plasticity operation is initiated. 
As in Eq. (4), $C$ is a constant bias term to account for negative error because traces cannot be negative. 

\section{Experiment}
\subsection{Setup}
\begin{figure*}
    \includegraphics[width=1.0\textwidth]{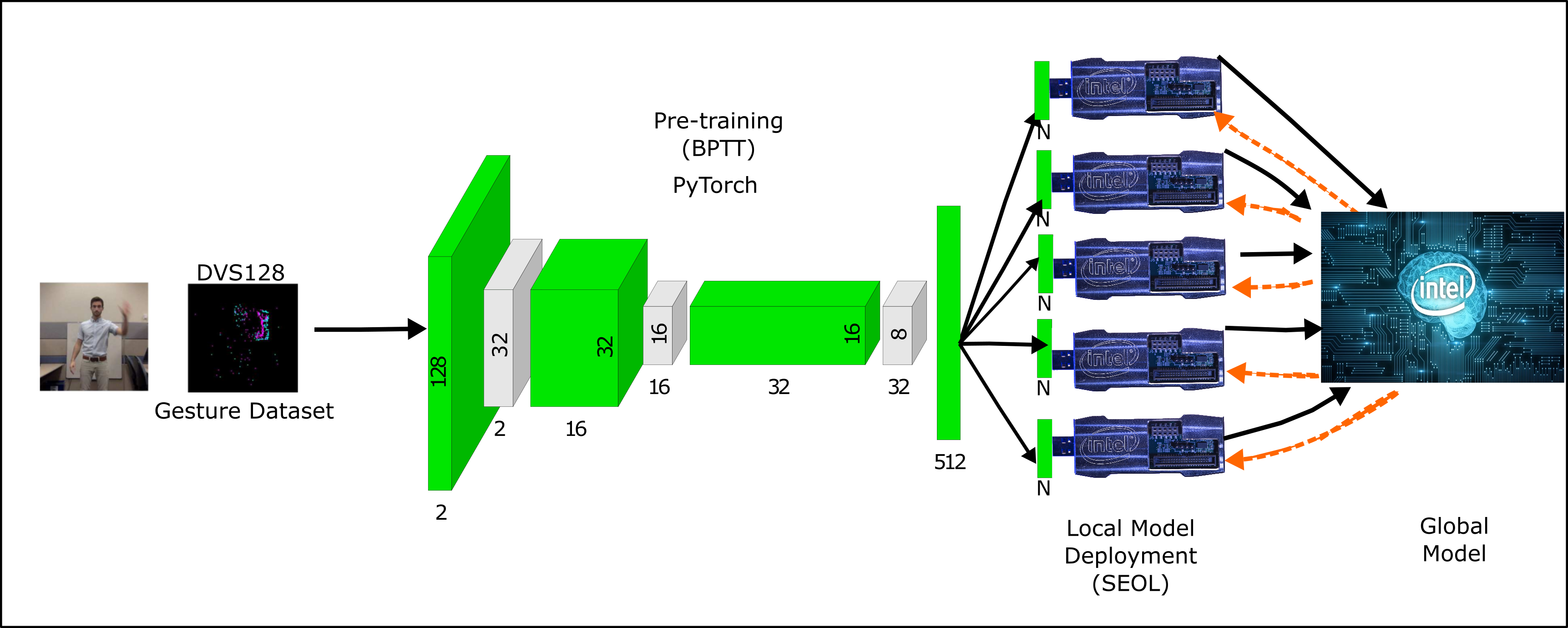}
    \centering
    \caption{Experimental setup. During a pre-training phase, the Loihi compatible convolutional network is trained on a GPU using DVSGesture dataset recorded using a DVS128 camera, the functional simulator, and SLAYER BPTT. The entire network along with quantized parameters of the functional simulation are then transferred on to the Loihi cores. In each model, one-shot learning is performed on the final layer using on-chip SOEL. The deployed network, including inference and training dynamics are performed on the Loihi chips. Dashed orange arrows indicate the transfer of averaged weights from the global models to the local models.}
    \label{fig:learning}
\end{figure*}

For the one-shot federated learning experiments we used the IBM DvsGesture dataset \cite{Amir_etal17_lowpowe}. The dataset consists of recordings of 29 different individuals performing 10 different actions such as clapping and an unspecified gesture for a total of 11 classes. The actions are recorded  using a Dynamic Vision Sensor (DVS) \cite{Lichtsteiner_Delbruck05_64x6aer}, an event-based neuromorphic sensor, under three different lighting conditions. The problem is to classify an action sequence video. Samples from the first 23 subjects were used for training and the last 6 subjects were used for testing. The test set contains 264 samples and will be used for the few-shot federated learning task. Each sample consists of the first 1.45 seconds of the gesture performed. 
Previous work has demonstrated the success of using transfer learning to boost the performance of few-shot learning accuracy using surrogate gradient learning methods like SOEL \cite{stewart2019onchip} \cite{stewart2020online}. 
In our experiment we use this approach with federated learning on an M+N-way classification task.
Transfer learning has the advantage that pre-training can be accelerated offline if the task domain is known, and few samples of each class are sufficient for learning the target task at reasonable accuracies. 
Therefore we used a network pre-trained using SLAYER \cite{Shrestha_Orchard18_slayspik}, an SNN learning method based on Back-propagation Trough Time (BPTT). The SLAYER model was pre-trained on only 6 of the 11 classes.
Then we transferred the network to the Intel Loihi and retrained the last layer of the network on the remaining 5 classes using the Neuromorphic Federated Learning (NFL) algorithm expressed in Algorithm 2, combining SOEL with federated learning that we use for 6+5-way one-shot federated learning.

\begin{table}[!ht]
\label{tab:arch}
\begin{center}
\caption{Network architecture.}
\def\x{$\times$}
\begin{tabular}{|c|c|c||c|} \hline
Layer & Kernel & Output    & Training Method\\ \hline
input &      & 128\x128\x2 & DVS128 (Sensor)\\\cline{4-4}
1     & 4a     & 32\x32\x2   & \multirow{6}{*}{SLAYER (BPTT)}\\
2     & 16c5z   & 32\x32\x16  &\\
3     & 2a     & 16\x16\x16  &\\
4     & 32c3z   & 16\x16\x32  &\\
5     & 2a     & 8\x8\x32    &\\
6     & -      & 512       &\\ \cline{4-4}
output   & -      & N = 5        & SOEL\\ \hline
\end{tabular}
% \undef\x
\end{center}
\scriptsize{Notation: \verb~Ya~ represents \verb~YxY~ sum pooling, \verb$XcYz$ represents X convolution filters (\verb~YxY~) with zero padding. $N$ is the number of classes, which is task-dependent. In our experiments N=5 because we are training on 5 classes.}
\end{table} 

Our experiment used five client models and one server, or global, model. 
The client and global models are SNN models trained on the Intel Loihi for the 6+5-way one-shot federated learning are described in Table I.
Hidden layers 1 through 6 are spiking CNN layers that were trained offline using SLAYER \cite{Shrestha_Orchard18_slayspik} on the Loihi neuron model. 
The output layer synapse weights are trained using SOEL.
Each client used one shot of the training data from the five classes of the DVSGesture dataset not pretrained using SLAYER. The clients trained their last layer using the SOEL learning algorithm.
After each client trained on their one-shot of data for an epoch the weights of each model are averaged by the global model.
Additionally, because the Loihi weights can only be 8-bit even integers the averaged weights were rounded to their nearest even whole number value. 
The averaged weights from the global model are then used as the new weights for each of the five client models for training again on their respective data. The process repeats for a number of epochs until the client models converge. 

\begin{algorithm}
\begin{algorithmic}
   \State \textbf{Server}:
   \State Server initializes common model $w_0$    
    \For {$t = 1$, $E$}   
        \For {each client $k$ in $K$}   
            \State $w_{t} = w_{t-1} +$\textbf{Client}$_k(w_{t-1}^k)$   
        \EndFor   
        \State $w_{t} = w_{t}/K$   
    \EndFor   
    
    \State \textbf{Output}: Global model $w_t$
    \\
    \State \textbf{Client$_k(w_{t-1}^k)$:}
    \State $W_{e}^{k} =$ SOEL$(D_k,w_{t-1}^{k},T)$
    \State \textbf{Output}: Model update $w_t^k - w_{t-1}^k$

\end{algorithmic}

\label{alg:NFL}
\caption{Neuromorphic Federated Learning}
\end{algorithm}

\subsection{Results}
\begin{table}[h!]
\centering
\caption{\label{tab:dvs_table5-way}6+5-way one-shot federated classification on the DvsGesture dataset 
}
\begin{tabular}{|l|r|r|}
\hline
Model & \multicolumn{1}{l|}{Initial Test Accuracy} & \multicolumn{1}{l|}{Post Federated Learning Accuracy} \\ \hline
1 & 76\%   & 88\%    \\ \hline
2 & 75.5\% & 88\% \\ \hline
3 & 58\%   & 85\%    \\ \hline
4 & 54.4\% & 86\%    \\ \hline
5 & 59.5\% & 81\%    \\ \hline
\end{tabular}
\end{table}

In Table \ref{tab:dvs_table5-way} we compare the test accuracy of doing 6+5-way one-shot learning of the five different client models and compare to the accuracy of each model after federated learning converged which in this case was after 8 epochs. All of the models improved on the test data after training on only one-shot of data using federated learning, with model 4's accuracy improving by almost 32\%. Models 1, and 2 achieve state of the art one shot learning accuracy on the gesture recognition task with the previous state of the art being done with SLAYER at 86\% \cite{stewart2020online}. The higher accuracy of the models are the result of learning for more than a single epoch, and better generalization from averaging the weights of the different models in the global model thus sharing some of the high level features learned from each individual model's shot. 

\section{Discussion and Future Work}
We achieved state of the art one-shot learning accuracy on a gesture recognition task by combining surrogate gradient learning with federated learning on a neuromorphic processor. 
These initial results are promising and will help pave the way for more secure, better performing mobile systems that use neuromorphic computing.
However, there are still many open problems remaining in making neuromorphic learning algorithms more secure.
We discuss these in the following sections and will address these challenges in future work.
\subsection{Security of Federated Learning}
As a relatively new concept, although Federated Learning creates new opportunities, there are still a few drawbacks remaining to be solved. Researchers have made intensive investigation into the security of the federated learning framework under different kinds of attacks \cite{fedopenproblems}. First, in practice there might be parties that don't follow the Federated Averaging algorithm and manipulate the model updates between clients and the server, which can degrade the overall model performance. For example, a malicious party could intentionally send bad model updates for aggregation which will potentially degrade the model quality. Moreover, this kind of attacks might introduce backdoor attacks \cite{fedbackdoor1} \cite{fedbackdoor2}. Second, although not directly uploaded, the private information about the training data of honest parties could still be extracted from their model updates by carefully designed attacks from another malicious party, especially when the number of parties participating in training process is small, e.g. only two clients participated. Although this could happen with negligible possibility in practice, we still want to prevent this kind of attacks theoretically and practically. Third, large models may require huge bandwidth during communication between clients and the server, since nowadays it is very common to have models with thousands or millions of parameters, the size of each update can be quite large and thus expensive.
\subsection{Other privacy-enhancing techniques}
Differential privacy is commonly used in deep learning to protect model privacy \cite{abadi2016differential}. In general it allows a party to privately release information about a dataset while an evaluation function on this input dataset is perturbed. When applied to machine learning models, carefully designed noise such as Guassian/Laplacian noise are added to gradients so that the information that could be learned from this gradient update is bounded.  
However noise permutation mechanisms are yet to be explored with surrogate gradients. Usually it's expected that adding noise would hurt model's performance, while our experiment shows that using federated learning could improve the experiment result which could balance out the performance degradation from the added noise.

Homomorphic Encryption is another method that could prevent private information leakage during the machine learning process. In most cases it uses a centralized learning setting, which generally enables the server to perform learning tasks on encrypted data uploaded by clients, and sends back to the clients the learned results, e.g. the predicted result output by a prediction model. During the whole process only the client can see its own raw data and corresponding learning result after local decryption. The whole machine learning model could be seen as a computation function, which runs on encrypted training data. 
However the computation cost of homomorphic encryption is very high, although recent work has proposed some schemes specifically for decreasing the computation cost \cite{ckks} \cite{tfhe}. Because the model used in our work and the models typically used in mobile and embedded applications are relatively small, homomorphic encryption to protect user data privacy while still expensive is worth exploring.

\section*{Acknowledgments}
This research was supported by the Intel Corporation.

\bibliographystyle{IEEEtran}
\bibliography{biblio_unique_alt,bib,apbib}

\end{document}